\documentclass[HARVARD,Times2COL]{WileyNJDv5}

\articletype{Article Type}%

\received{8 November 2024}
\revised{10 September 2025}
\accepted{24 October 2025}
\journal{Expert Systems}
\volume{00}
\copyyear{2025}
\startpage{1}
\doi{10.1111/exsy.70164}

\begin{document}

\title{The challenge of generating and evolving real-life like synthetic test data without accessing real-world raw data - a Systematic Review}

\author[1]{Maj-Annika Tammisto}

\author[2]{Faiz Ali Shah}

\author[3]{Daniel Rodriguez}

\author[4]{Dietmar Pfahl}

\authormark{Tammisto \textsc{et al.}}
\titlemark{Generating and evolving real-life like synthetic data - a systematic review}

\address[1]{\orgdiv{Institute of Computer Science}, \orgname{University of Tartu}, \orgaddress{\state{Tartu}, \country{Estonia}}, \email{maj-annika.tammisto@ut.ee}}

\address[2]{\orgdiv{Institute of Computer Science}, \orgname{University of Tartu}, \orgaddress{\state{Tartu}, \country{Estonia}}}

\address[3]{\orgdiv{Department of Computer Science}, \orgname{University of Alcala}, \orgaddress{\state{Madrid}, \country{Spain}}}

\address[4]{\orgdiv{Institute of Computer Science}, \orgname{University of Tartu}, \orgaddress{\state{Tartu}, \country{Estonia}}}






\abstract[Abstract]{
\textbf{Background:} High-level system testing of applications that use data from e-Government services as input requires test data that is real-life-like but where the privacy of personal information is guaranteed. Applications with such strong requirement include information exchange between countries, medicine, banking, etc. This review aims to synthesize the current state-of-the-practice in this domain.

\textbf{Objectives:} The objective of this Systematic Review is to identify existing approaches for creating and evolving synthetic test data without using real-life raw data.

\textbf{Methods:} We followed well-known methodologies for conducting systematic literature reviews, including the ones from Kitchenham as well as guidelines for analysing the limitations of our review and its threats to validity. 

\textbf{Results:}  A variety of methods and tools exist for creating privacy-preserving test data. Our search found 1,013 publications in IEEE Xplore, ACM Digital Library, and SCOPUS. We extracted data from 75 of those publications and identified 37 approaches that answer our research question partly. A common prerequisite for using these methods and tools is direct access to real-life data for data anonymization or synthetic test data generation. Nine existing synthetic test data generation approaches were identified that were closest to answering our research question. Nevertheless, further work would be needed to add the ability to evolve synthetic test data to the existing approaches. 

\textbf{Conclusions:} None of the publications really covered our requirements completely, only partially. Synthetic test data evolution is a field that has not received much attention from researchers but needs to be explored in Digital Government Solutions, especially since new legal regulations are being placed in force in many countries.
}

\keywords{Synthetic Test Data, Data Synthesis, Synthetic Data Generation, Artificial Data, Synthetically Generated Data, Data Evolution}

\jnlcitation{\cname{%
\author{Tammisto M-A.},
\author{Shah F A},
\author{Rodriguez D}, and
\author{Pfahl D}}.
\ctitle{Generating and evolving real-life like synthetic test data - a systematic review} \cjournal{\it Expert Systems} \cvol{202x;00(00):1--18}.}

\maketitle

\renewcommand\thefootnote{}
\footnotetext{\textbf{Abbreviations:} DGS, Digital Government Solutions; SUT, System Under Test; QA, Quality Assessment; GDPR General Data Protection Regulation 
}

\renewcommand\thefootnote{}
\footnotetext{\textbf{Data availability:} \url{https://doi.org/10.6084/m9.figshare.25809616.v1}
}

\renewcommand\thefootnote{\fnsymbol{footnote}}
\setcounter{footnote}{1}

\section{Introduction}
\label{sec:intro}

Digital Government Solutions (DGS), also referred to as e-Government solutions, are designed to provide public services without using extensive manpower and bureaucracy. These services cover a wide variety of applications such as taxes, utility bills, licenses and permits, medical information, post service for official documentation, etc. They enable the general public to communicate with the government conveniently and efficiently and are implemented and used in most of the countries in the world on a smaller or larger scale.

However, there are limitations when it comes to using the actual data that is processed by government entities for activities that are not part of the actual e-Government service provision, such as pre-production testing. In Europe, a large portion of this data is considered to be personal data according to the General Data Protection Regulation (GDPR)~\citep{GDPR}. The United States follows a sectoral approach to data privacy protection~\citep{Boyne18} and the growth of new digital industries has motivated Asian countries, including China, to work on their legislation in order to restrict the use of personal data~\citep{Junke21}. These regulations make actual personal data unavailable for testing and force Quality Assurance Specialists all over the world to find solutions for creating or obtaining privacy-preserving test data that is as similar as possible to the actual data processed by the government.

A variety of methods and tools exist for creating privacy-preserving test data. One practical solution is data anonymization which transforms the original data by applying some operations on it to effectively remove actual personal data without degrading the anonymous data utility~\citep{Majeed21}. Nevertheless, there remains the risk of someone reversing the anonymization algorithm and retrieving actual personal data. Another one of the many possible options would be to use one of the various existing machine learning models and generate fully synthetic test data that is very similar to the actual data, provided that the model is trained well. A common prerequisite for using most of these well-known options is that they require direct access to the actual real-life raw data that is used as input for data anonymization or synthetic test data generation. However, failure to gain access to actual government raw data excludes the possibility of using the above-mentioned methods.

Another aspect that needs to be considered when creating privacy-preserving test data that is as similar as possible to actual data processed by governments is the fact that actual data is constantly evolving. This is an important consideration to note because although historical data from months or even years ago that contains events with timestamps, as well as consistent relations between data subjects, is sometimes important for providing e-Government services, the majority of applications rely on data that reflect the current or the recent state of the data subject. For example, banks may query income information about the latest months in order to calculate credit limits. Some family benefits might be granted only to parents with newborns under a certain age. Some applications may require a recent life event (e.g. birth, marriage, divorce, or death) as input. Therefore, one could say that a static set of test data created for e-Government entities has an ``expiration date'', as over time it will become more and more useless for the applications under test. For that reason, it is important that the test data resembling actual data processed by the government is created with knowledge of the evaluation mechanisms of real-life data and that the test data can be evolved in a similar manner.

One possible solution would be to generate and evolve synthetic test data based on publicly available microdata, open data, or other input that captures essential characteristics and distribution of the actual real-life raw data to be mimicked without revealing any personal information. The objective of this study is to identify and describe existing methods in the field of Software Testing that are able to do that.

The rest of this paper is structured as follows. Section~\ref{sec:background} summarizes the background of this study, including our specific context. Section~\ref{sec:relatedWork} describes related work. Section~\ref{sec:methodology} provides definitions and explains our research method. In Section~\ref{sec:results}, the results of our study are presented. Section~\ref{sec:ttv} gives an overview of the limitations of this study, as well as the resulting threats to validity. Section~\ref{sec:discussion} provides the discussion of results, and finally, Section~\ref{sec:conclusions} concludes the paper and highlights possible future research directions.

\section{Background}
\label{sec:background}

Estonia, one of the leading countries in e-Government development according to the United Nations E-Government survey of 2022~\citep{EGov22}, is one of the pioneers in implementing digital government solutions. Estonian e-Government solutions are built on the interoperability framework X-Road\footnote{\url{https://x-road.global/}}. Today, other countries, for example, Finland, Iceland, and the Faroe Islands have also implemented the X-Road solution and as a next step, the first cross-border data exchange project was started between Finland and Estonia~\citep{Jackson21}.

\begin{figure*}
    \centering
    \includegraphics[width=1.0\linewidth]{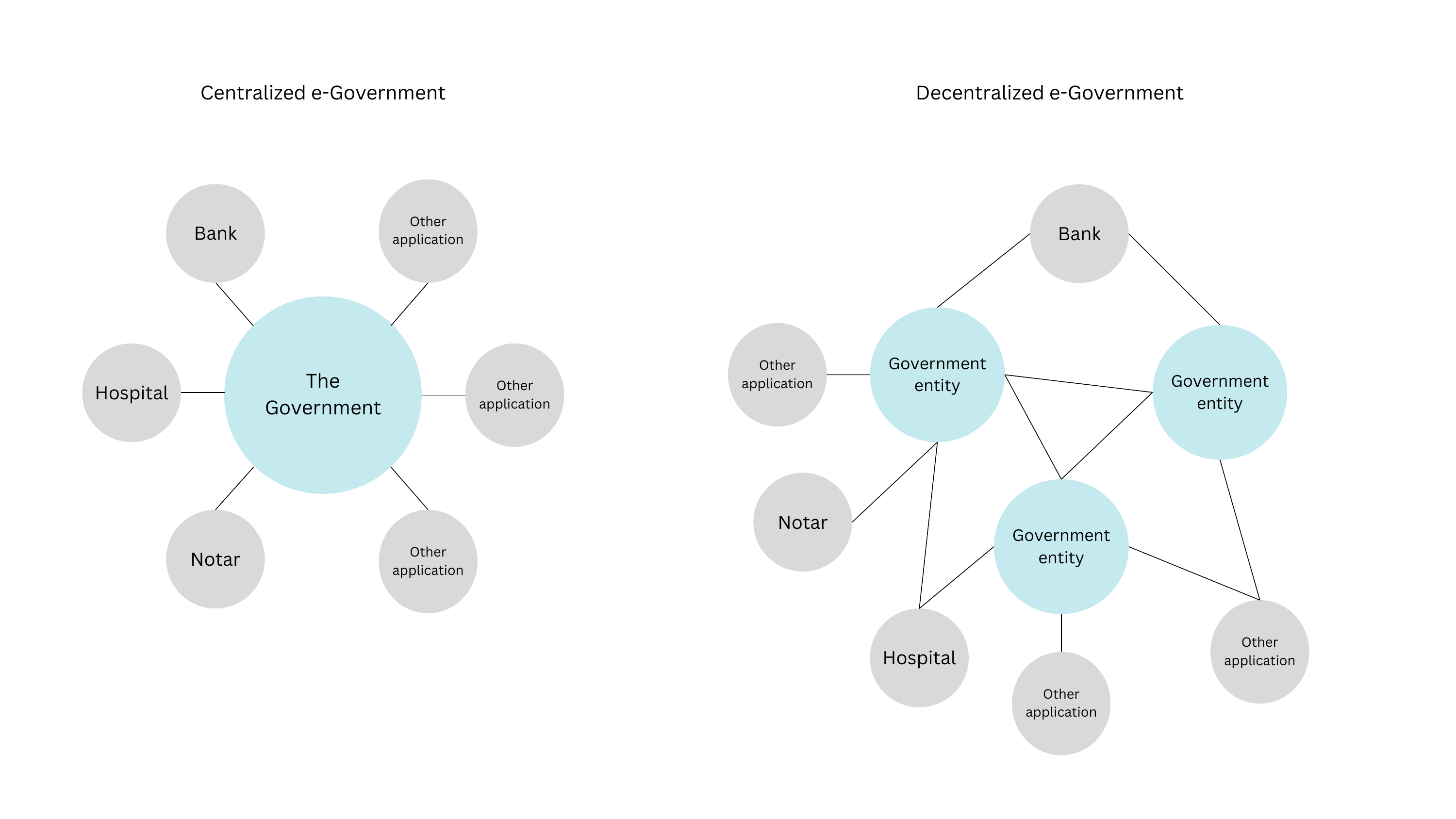}
    \caption{Centralized vs. Decentralized DGS}
    \label{fig:cVdDSG}
\end{figure*}

For several decades, many countries have pursued decentralization of government services with the objective of improving service delivery \citep{Gradstein17}. In decentralized DGS settings, there is no central database that can be queried for all government data. It is a network of government entities that act as data providers who are able to exchange data with other parties (see Figure~\ref{fig:cVdDSG}). Driven by the principle ''Data resides where it is created'', Estonian e-Government falls into the category of decentralized e-Government, as there is no central database that can be queried via X-Road for all data that the government stores~\citep{Veldre16}. Instead, different government institutions create data during the execution of their procedures and operations, they are also responsible for maintaining their data, and where justified, they enable public and private sectors to connect with them via X-Road data services and to request their data. 

Interoperability is the basis of a decentralized DGS. Important factors for designing an underlying interoperability framework for e-Government services  are well studied \citep{Flak12,Scholl07}. The challenges related to testing the applications that are using the interoperability framework and the data from decentralized e-Government services as input have on the other hand received little attention from researchers and industry.

When actual data is required in the case of centralized DGS, an appropriate option for creating privacy-preserving test data would need input from one central government database, however, for decentralized DGS this would mean having access to the data of all separate government entities. It is a challenge that requires compliance with the individual security procedures of every government entity. Therefore avoiding directly accessing actual government data when creating test data would reduce the risk of data breach as well as the procedural complexity significantly.

This multi-source data received is thereafter used as input by the public or private sector parties that requested it, in their digitalized work processes such as providing (public) services, monitoring, reporting, or similar. Given that there is no central government database in a decentralized data exchange framework like the X-Road, there is also no central development of e-Government services as they are planned, developed, and implemented independently by separate government institutions. The increasing number of data services that new e-Government developments use for input poses new challenges to pre-production testing as these applications need to be tested thoroughly without the actual production data being available for testing due to data privacy restrictions.

Today, there is no test data in Estonia that is dedicated to testing applications that use X-Road data service responses as input. The test data that is available for pre-production testing is created by government institutions in their pre-production environments mainly as a result of testing their own applications and mostly without considering that they are also acting as test data providers. That means that currently available input data for testing new e-Government services is often incomplete or even not possible in real life. Considering that pre-production environments of different government institutions are not synchronized, test data received from more than one institution is on many occasions not compatible when used as input for testing. It is possible to order specific test data from government institutions via e-mail, but it is a time-consuming and expensive process with a high risk for human error as test data is in most cases created manually, either through the user interface, by manipulating the pre-production database or by combining
these two actions

The obvious solution for overcoming the challenges of testing with Estonian X-Road data services would be to oblige all X-Road members who provide data services to obey a specific set of rules that applies to the quality and availability of test data provided by them via X-Road in pre-production environments. Considering that there is to date no common Service Level Agreement in place for pre-production environments of Estonian X-Road members, this needs to be a strategic decision made by the government level and backed up with sufficient resources for implementing this change. An alternative solution would be to create one centrally managed dataset that is dedicated to providing test data for data services in X-Road pre-production environments. A significant amount of data moving through X-Road in production is considered sensitive data in terms of the General Data Protection Regulation (GDPR)~\cite{GDPR}. There are profound consequences for all parties if such case sensitive data were to become publicly available. To mitigate any risks of data breaches, the centrally managed dataset cannot contain any sensitive data or any anonymized versions of that data. This means that the centrally managed dataset must be fully synthetic.

Any methods to be considered for generating the synthetic dataset need to be suitable for generating synthetic data based on the characteristics of not only one database but several independent datasets, hereinafter referred to as multi-source data. The growing number of projects where proactive services are developed indicates that the synthetic dataset cannot be a static one. Proactive e-Government services use certain life events that are reflected in data as triggers for proactively offering services and benefits to citizens and residents instead of citizens and residents applying for these services and benefits. Therefore, any test data used as input for testing proactive e-Government services should have an evolution logic that is similar to real-life data that is used in production.

\begin{figure}
    \centering
    \includegraphics[width=\columnwidth]{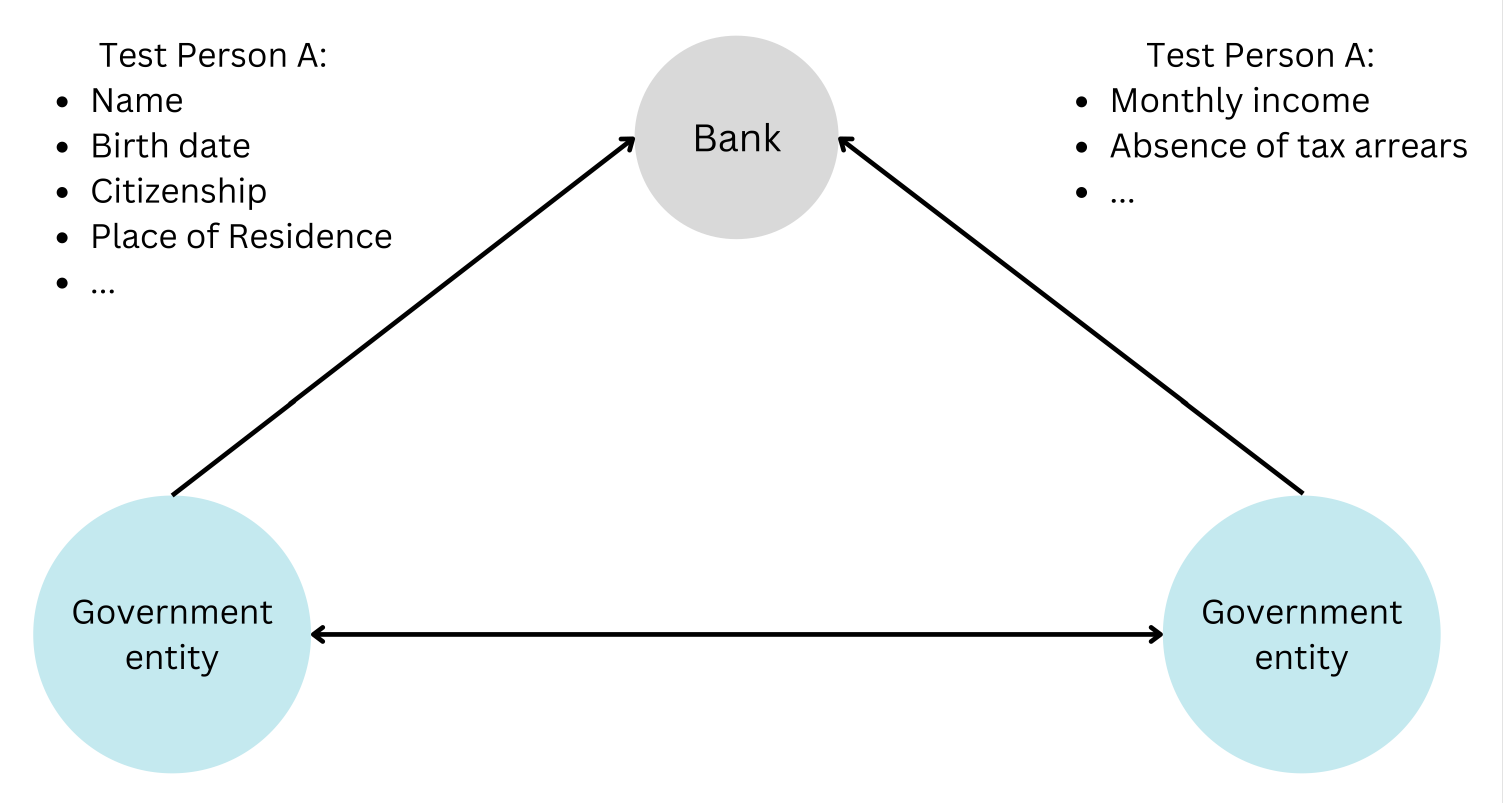}
    \caption{Compatibility of test data among different government entities in a decentralized DGS}
    \label{fig:compTest}
\end{figure}

In a decentralized DGS it is necessary to create test data in a way that the test data of every government entity would be compatible with the test data of every other government entity that is part of the same e-Government solution. More specifically, there are certain identificators that must be preserved when creating the test data created for every government entity in a decentralized e-Government solution. This is crucial for ensuring that all test data in the decentralized DGS is consistent and an instance of test data provided by one government entity (for example a test person) has a valid and meaningful match among the test data of another government entity as (see Figure~\ref{fig:compTest}).

The growing number of projects where proactive services are developed indicates that the test dataset cannot be a static one. Proactive e-Government services use certain life events that are reflected in data as triggers for proactively offering services and benefits to citizens and residents instead of being the citizens and residents applying for such services and benefits. Therefore, any test data used as input for testing proactive e-Government services should have an evolution logic that is very similar to real-life data that is used in production.

\section{Related work}
\label{sec:relatedWork}

Due to the need to protect personal data or the lack of real-life data, there is an active area of research looking for effective methods for generating synthetic data. It is not only the discipline of software testing that requires data that closely resembles real-life data, but where all personal data that may lead to the identification of an individual is removed.

The development of Machine Learning (ML) algorithms relies on training data and is challenged by data privacy requirements as stated by~\cite{Abufadda21}, which have reviewed a number of the models and studies that proposed generating or using synthetic data for ML in various medical, scientific and social fields. This study lists several synthetic data generation approaches but provides only a few insights in every approach listed. There is no evidence that the approaches that they identified are able to generate and evolve synthetic data without using real-life data as input. 

They did not analyse the current trend of LLMs (Large Language Models) for this task, nor did we find relevant approaches to this technology in our domain, which we intend to explore as it is discussed later in this review.

The survey carried out by 
~\cite{Eigenschink23}
evaluates deep generative models for synthetic sequential data based on their representativeness, novelty, realism, diversity and coherence. The authors concentrate on assessing the similarity of the generated synthetic data to actual real-life data, which is also very important in our context. However, another aspect that is relevant for our context, the total avoidance of using actual real-life data, was not considered in this study.

Also, 
\citet{Gobel23} have addressed the large gap between publicly available datasets and actual needs in the field of digital forensics and provided a list of available dataset generation frameworks. Again, the emphasis of this study is not on generating synthetic data without using any real-life data in the process. 

Our study fills this gap and concentrates on synthetic data generation approaches that do not use real-life data as input.


\section{Research methodology}
\label{sec:methodology}

Our research team consists of four researchers. We are using Kitchenham's guidelines for performing Systematic Literature Reviews in software engineering  for selecting relevant publications~\citep{Kitchenham07}.


\subsection{Definitions}\label{subsec4:definitions}

In the following, we list definitions of three concepts as we understand them in the context of our study.

\begin{itemize}
\item \textbf{Real-life raw data:} This refers to data that is created, gathered and processed in real-world settings and that is not publicly available.
\item \textbf{Synthetic test data:} This refers to artificially created test data that can be used to replace real-life raw data in high-level system testing.
\item \textbf{Evolving synthetic test data:} This refers to transforming the generated synthetic test data over time and in doing so preserving a set of essential attributes of data object instances, e.g., the relationships between two or more data subject instances.
\end{itemize}


\subsection{Research Question}
\label{subsec:rq}

To meet the objective of this study, the research question (RQ) is:

{\bfseries What methods exist for generating and evolving synthetic test data that imitate real-life data without using the respective real-life raw data as input?}

The aspects of interest related to our RQ are:

\begin{itemize}
\item Type and characteristics of input data used for synthetic test data generation.
\item Description of synthetic test data generated.
\item Data evolution ability of synthetic test data generation methods.
\end{itemize}

We are particularly interested in exploring test data generation methods that do not require real-life raw data, yet can generate synthetic test data that closely resembles real-life data. In addition, we aim to generate synthetic test data that evolves in a manner closely resembling the evolution of real-life data. One approach to achieve this is to repeatedly generate new synthetic test data along a sequence of time steps, where the test data generated at each subsequent time step uses the data from the previous time step as input. Alternatively, there may exist methods that do not require the repeated generation of test data, but instead evolve the initially generated test data, similar to dynamic simulation. Therefore, we are also interested in exploring methods that allow synthetic test data to evolve over time.

\subsection{Search for Publications}

Inspired by the guidelines of \cite{Kitchenham07}, the research question is broken down into individual facets (also considering the PICo criteria\footnote{\url{https://utica.libguides.com/c.php?g=960363&p=6934097}}  for qualitative research used for defining the facets, after which a list of synonyms and alternative spellings is created). Before conducting the actual search, trial research strings are created and tested against a list of already known primary studies. The research question facets that are defined as too restrictive while testing trial research strings were removed from the final research strings and included as conditions in the inclusion and exclusion criteria.

Additionally, the list of synonyms is fine-tuned with the help of \texttt{wordfreq}\footnote{\url{https://pypi.org/project/wordfreq/}}\citep{Speer22}, a Python library for looking up the frequencies of words in many languages, based on many sources of data. This library defines the most frequent keywords describing the Problem, Phenomenon of interest, and Context used in reference articles and it helps the authors to define and include the most important keywords in the search strings.

The following search strings were used:

\begin{itemize}
\item \textbf{Search string used on IEEE Xplore:}
(Advanced search -> Command search (Boolean/Phrase): \textit{("software test*" OR "software quality" OR "quality control" OR "quality assurance") AND ("synthetic data*" OR "data synthesis" OR "artificial data" OR "synthetically generated data" OR “random data generation”)})
\item \textbf{Search string used on ACM Digital Library:}
(Advanced search -> The ACM Full-Text collection -> Search within Anywhere): \textit{("software test*" OR "software quality" OR "quality control" OR "quality assurance") AND ("synthetic data*" OR "data synthesis" OR "artificial data" OR "synthetically generated data" OR “random data generation”)})
\item \textbf{Search string used on Scopus:}
(Search -> Refine search -> Subject area: limit to Computer Science): \textit{("software test*" OR "software quality" OR "quality control" OR "quality assurance") AND ("synthetic data*" OR "data synthesis" OR "artificial data" OR "synthetically generated data" OR “random data generation”)})
\end{itemize}

After conducting the database search for RQ, all publications found with the search are immediately exported to the Zotero reference management tool\footnote{\url{https://www.zotero.org/}} where every search result receives a unique ID. For further analysis, the results are exported from Zotero to a shared spreadsheet database that is used as the main working document by all four researchers.

All duplicates are identified and removed before proceeding with the Title and Abstract Analysis.


\subsection{Title and Abstract Analysis}

The Title and Abstract Analysis stage consists of the following two steps.

\begin{itemize}
\item \textbf{Filtering of Publications based on Exclusion Criteria:}
Exclusion Criteria are applied to every unique publication
found with the search. The purpose of applying the Exclusion Criteria first is to efficiently exclude publications that cannot be included for Full Text Analysis. 
\item \textbf{Inclusion of Publications for Full Text Analysis:}
in the Inclusion for Full Text Analysis step, only the Title and
Abstract of publications are read and analyzed with regard to our two Inclusion Criteria.
To be included for Full Text Analysis, the publication has to be a primary study that meets both of our two Inclusion
Criteria. 
\end{itemize}

Every publication included for Full Text Analysis is given a unique ID (P for Publication + number).

The Exclusion and Inclusion Criteria used are provided in Tables~\ref{tab:exclusion} and ~\ref{tab:inclusion}.

\begin{table*}
  \caption{Exclusion Criteria}
  \label{tab:exclusion}
    \begin{tabular*}{\textwidth}{@{\extracolsep\fill}lp{17cm}@{\extracolsep\fill}}
    \toprule
    \textbf{ID} & \textbf{Exclusion Criteria}\\
    \midrule
    E1 & Book, book section, or a conference review. 
    \textit{Justification:  this study aims to define approaches that are described with enough detail and quality that they are published in research papers.}\\
    \hline
    E2 & Full text is not available. 
    \textit{Justification: it is not possible to extract the data necessary for our study from a publication that is not available in full. By “not available” we mean “it cannot be accessed under our existing licenses, and it cannot also be purchased separately online”.}\\
    \hline
    E3 & Full text is not available in English.
    \textit{Justification:  although translation services and software are available for most languages worldwide, we cannot be certain that all technical details are presented correctly if the authors themselves do not translate the paper.} \\
    \bottomrule
  \end{tabular*}
\end{table*}

\begin{table*}
  \caption{Inclusion Criteria}
  \label{tab:inclusion}
  \begin{tabular*}{\textwidth}{@{\extracolsep\fill}lp{17cm}@{\extracolsep\fill}}
    \toprule
    \textbf{ID} & \textbf{Inclusion Criteria}\\
    \midrule
    I1 & The publication must mainly suggest and describe an approach or approaches for real-life-like synthetic test data generation.
    
    \textit{Justification:  papers that are not mainly concentrated on real-life-like synthetic test data generation are not likely to provide us with enough information to be able to use the approach in our next study.}\\
    \hline
    I2 & No real-life data must be required as input or training data in any step of the test data synthesis process.  
    
    \textit{Justification: approaches that use statistics, publicly available metadata, or any other means that do not require direct access to actual real-life data are to be defined.}\\
    \bottomrule
  \end{tabular*}
\end{table*}


\subsection{Full Text Analysis}

In turn, the Full Text Analysis stage is also composed of the following two steps.

\begin{itemize}
\item \textbf{Exclusion of Publications:}
In case there are secondary studies that were not identified 
in the previous stage, they will be identified and removed from further analysis. The remaining publications are re-assessed based on our Exclusion and Inclusion criteria to define those where proper exclusion was not possible based on the Title and Abstract only. If a publication gets excluded in this step, the prefix of the unique ID is changed from 'P' to 'Ex'. All publications that are not excluded in this step are selected for further analysis.
\item \textbf{Data Extraction:}
Conceptually, data extraction from the included publications focuses on data extraction items related to (i) study description, (ii) answers related to our RQ, (iii) quality of the publication, and (iv) maturity of the proposed approach. 
\end{itemize}

The Quality Criteria used for assessing the quality of the publication (iii) are provided in Appendix~\ref{app2}.

\subsection{Data Synthesis}

To find an appropriate method for synthesizing our extracted data, we have looked in the toolbox of Qualitative Data Synthesis (QDS) which is based on identifying common themes across qualitative studies to create a great degree of conceptual development compared with narrative reviews~\citep{Hollier}. One example of QDS is Thematic Synthesis, a straightforward method with clearly described steps~\citep{Flemming21,Thomas08} where data synthesis is traditionally carried out in several stages. The following stages are used in our study: 

\begin{enumerate}
    \item \textbf{Line-by-line coding of the findings of the individual studies:} a mixed approach is used as the data extraction sheet includes some classifications that are defined based on our domain knowledge. Additional coding is done for extracted data items that contain free text.
    \item \textbf{Development of descriptive themes:} reviewers group the created codes into a hierarchical tree structure based on code similarities and differences.
\end{enumerate}


\section{Results}\label{sec:results}


\subsection{Results of the Search for Publications}

Our search across the three selected databases produced a list of 1,013 publications. Table~\ref{tab:DLsUsed} shows the number of publications found in each of the digital libraries used.


\begin{table}
\caption{Digital Libraries used}
\label{tab:DLsUsed}
\begin{center}
\begin{tabular*}{250pt}{@{\extracolsep\fill}lcc@{\extracolsep\fill}}%
\toprule
\textbf{Digital Library} & \textbf{URL} & \textbf{\# of papers} \\
\midrule
IEEE Explore & \url{https://ieeexplore.ieee.org/} & 210 \\
ACM DL & \url{https://dl.acm.org/} & 402 \\
SCOPUS & \url{https://www.scopus.com/} & 401 \\
\bottomrule
\end{tabular*}
\end{center}
\end{table}

All publications were imported into Zotero and organised into three separate folders. Among the full list of publications (1,013), 45 duplicates were identified and removed from further analysis.


\subsection{Results of the Title and Abstract Analysis}

The first step of the Title and Abstract Analysis involved pre-selecting publications and excluding those that did not meet our Exclusion Criteria for Full Text analysis. The following number of publications were excluded:

\begin{itemize}
    \item E1: 127 books or book sections excluded
    \item E2: 20 publications were excluded because the full text was not available
    \item E3: 2 publications were excluded because the full text was not available in English
\end{itemize}

After excluding 149 publications in the pre-selection stage, we continued our analysis with a list of 819 and proceeded with evaluating each publication based on our Inclusion Criteria. 
To include a publication for Full Text Analysis, it had to be a primary study where both of our two Inclusion Criteria had to be fulfilled.

We quickly discovered that I2 was often difficult to evaluate based on Title and Abstract only, therefore in order to not lose any relevant findings, we decided to include all findings where it was not clearly understandable from the Title and Abstract that real-life data is used as input for the suggested approach.

\begin{itemize}
    \item I1: 686 publications were not included because it was evident from the Title and Abstract that they do not suggest or describe an approach or approaches for generating synthetic test data that resembles real-life data.
    \item I2: 58 publications were not included because it was clear from the Title and Abstract that real-life data is required as input or training data in the test data synthesis process. 
\end{itemize}

As a result of the Title and Abstract analysis, 75 publications were included for Full Text Analysis and Data Extraction.

\subsection{Results of the Full Text Analysis}

After reading the full texts, it was clear that nine of the publications (Ex30, Ex38, Ex39, Ex40, Ex44, Ex45, Ex46, Ex54, Ex63) did not suggest any novel synthetic data generation approach as required in I1, therefore, they were excluded from the final selection. Additionally, another 29 of the included publications suggest a synthetic data generation approach where real-life data is used as input for generating synthetic data. Consequently, these 29 papers were also excluded based on I2.

\begin{figure*}[t]
\centerline{\includegraphics[width=\textwidth]{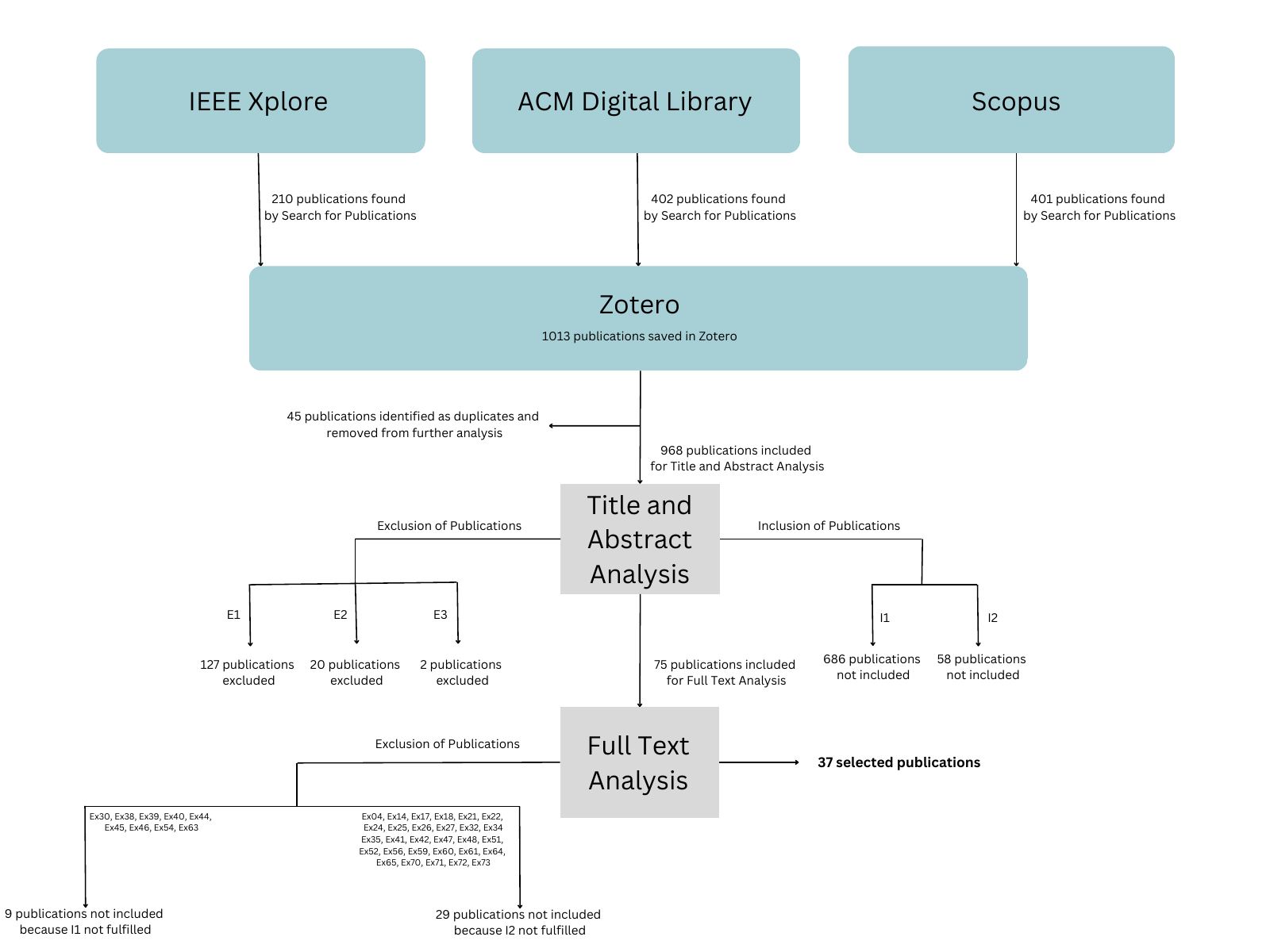}}
\caption{Results of the Title and Abstract Analysis and the Full Text Analysis}
\label{fig1}
\end{figure*}

Therefore, 38 publications in total were excluded in the Full Text Analysis stage, leaving 37 publications available for synthesis of the extracted data. The process followed for selecting these 37 publications is illustrated in Figure~\ref{fig1}. 

The demographics of these 37 selected publications are provided in Appendix~\ref{app3}.

\subsection{Answer to the RQ - Results of the Synthesis of Extracted Data}

We were prepared for the possibility of not finding a publication that provides us an answer for our whole RQ ``What methods exist for generating and evolving synthetic test data that imitate real-life data without using the respective real-life raw data as input?" Nevertheless, we were hoping that there were existing approaches that answer it partly. 

Therefore, we have split our RQ into individual RQ facets. The purpose is to look for answers with each one of these facets and to find out, which are the selected publications that answer the most of them. 

\subsubsection{What methods exist for generating synthetic test data?}

Our first Inclusion Criterion (I1) was designed as our ``line of defense" for not selecting publications that do not even suggest a novel synthetic data generation approach. Therefore, all of our 37 selected publications provide an answer to the RQ facet ``What methods exist for generating real-life-like synthetic test data?"

We have categorized the different types of approaches presented in the 37 selected studies as follows:

\begin{itemize}
    \item \textbf{Rule-Based generation:} approaches where synthetic test data is generated based on specific user-defined rules, or where the source code or the System Under Test (SUT) is used for defining the rules for synthetic test data generation.
    
    \item \textbf{Evolutionary Algorithms:} Algorithms that are based on the idea of evolution, for example Genetic Algorithms.
    
    \item \textbf{Classification/Regression Models:} Algorithms that generate synthetic test data based on previously trained classification/regression models.
    
    \item \textbf{Deep Learning:} Neural Networks with multiple layers of interconnected nodes, also referred to as neurons or units.
    
    \item \textbf{Image/Video Rendering Tools:} approaches where synthetic test data is created by using image and/or video rendering tools.
    
    \item \textbf{Simulation Environments:} approaches where synthetic test data is created by using simulation environments.
    
    \item \textbf{Other:} types of approaches that were suggested only once in our population of 37 publications.
\end{itemize}

There were a few cases where a suggested approach seemed to combine more than one method. For example, the approach suggested in P30 was classified as ``Rule-Based generation'', but there remains the question if the tool developed by the authors might also use a Search-Based algorithm, as meta-heuristics are used in the induction of the rules. In order to avoid unnecessary complexity in categorizing the types of approaches, we looked at this question from the viewpoint of the user of the approach and asked ourselves how the user of this approach would identify it. Would the user need to know the ``business logic'' to be able to create rules? Or understand how a search problem is solved? Or maybe have to have access to training data to train a model? The answers to these questions helped us identify the most relevant category from the user's point of view.

A total of 13 studies suggested a synthetic test data generation approach based on rules. This includes Web Services Description Language (WSDL) specifications (P01), information from specification/implementation or other formal requirements (P02, P07, P11, P19, P20, P53, P67, P69, P74, P75), tokens and grammar rules (P03), and UML diagrams and OCL constraints (P28). 

The category of Evolutionary Algorithms includes five suggested approaches. Genetic Algorithms (P06, P08, P09, P13) e.g. Search-based mutation testing (P05), are used in all cases.

We identified two cases where Classification/Regression Algorithms were used. A white-box regression model was suggested, where the structure of the model is available and can be used for test case generation (P12). Another publication (P62) offered a solution for classifying datasets that have great variability in the number of attributes, types of attributes, and number of class values.

The Deep Learning category includes five approaches where Deep Neural Networks (P29) or Generative Adversarial Networks (GANs) (P36, P49, P58, P66) are used for synthetic data generation.

Image/Video Rendering Tools were used in three studies. One publication (P15) used both commercial, and open-source software for synthesizing a 3D scene model with a city model and pedestrians and another publication (P55) combined two existing image rendering tools to synthetically generate facial data. In our third approach in this category, CAD models were combined with physical objects (P31) to generate data for manufacturing.

In the Simulation Environments category, we had an approach called SynTiSeD (P37) where simulation environments were used to synthesize energy consumption data. We also identified an approach SoccER (P57) that was built on the existing upon the Gameplay Football simulator and used to simulate football games with synthetic data. Our third selected publication in this category used simulation for creating datasets for the evaluation of Multi-Target Regression and Multi-Label Classification methods (P68).

The types of approaches that occurred only once among our 37 selected publications were assigned in the Other category. There are five approaches in total in this category, and they include:
\begin{itemize}
    \item A publication (P10) that aimed to evaluate the performance of hill climbing search algorithm compared to random test data generation in a very specific context
    \item A Successive Random Addition (SRA) method for creating synthetic weather patterns (P16)
    \item An approach called BackTranScription (BTS) that was suggested by the authors in their previous work and used for synthesizing speech in Korean (P33)
    \item A Domain-Specific Language (DSL) called Steveflex that was designed for creating synthetic data to test data-intensive software systems (P43)
    \item A generative hierarchical probabilistic dynamic model was proposed that explicitly models large spatial and temporal variations in human motion sequences (P50)
\end{itemize}

The purpose of generating synthetic test data with the suggested approaches was in roughly half of the selected publications (19) to facilitate software testing at different levels, from unit testing to high-level system testing. Other purposes included e.g. generation of training data for Deep Neural Networks, assessing or controlling the quality of certain domain-specific data, evaluating specific algorithms or models, system development in general, or research.

Specific limitations related to the suggested approaches were mentioned in 17 of the included publications. The limitations were concerning (i) the computational resources required for synthetic data generation, (ii) the performance of the suggested approach or tool, (iii) the limited types of output data that can be generated, and (iv) the quality of synthesised data. There are 20 studies that did not mention any limitations related to the suggested approaches.

\subsubsection{What methods exist for evolving synthetic test data?}

In order to answer this facet of our RQ, we first had to clearly define the meaning of ``evolving synthetic test data'' in our context. A definition is given in the Subsection~\ref{subsec4:definitions}, and it is important to note that it refers to the evolution of a set of synthetic data that is already created. This means, that in our context, data evolution cannot be a part of the synthetic data generation process. It is an independent process that can start only once the original synthetic test dataset is fully generated and its quality verified. Based on these thoughts, we were able to rule out Evolutionary Algorithms that were used for synthetic data generation as well as any other means of evolution that were not designed as an ongoing process for constantly evolving the synthetic test data.

Among our 37 selected publications, we identified only two studies that provided an answer to our RQ-facet ``What methods exist for evolving synthetic test data?" (see Figure~\ref{fig4}). Both studies are a part of the same research at the University of Oslo, Norway, in cooperation with Testify AS. The context of this research is in fact very similar to ours, as it was aimed at finding a solution for synthesizing and evolving test data that imitates the actual data in the Norwegian National Registry. 

The first one of the two publications (P66) was published in 2019 and suggests Multi-layer Recurrent Neural Networks for simulating life events (e.g. births, marriages, deaths) that are used to evolve the initial synthetic dataset that was created in the same publication. Although not specifically mentioned, it is likely that this approach is able to preserve essential attributes of data object instances. The training dataset that was used in this study was collected from a test environment of the Norwegian National Registry. The training dataset itself was therefore synthetic and generated in the course of several years either manually or through the execution of automated test suites. As a result, the statistical characteristics of this dataset were very different from those of the real Norwegian National Registry data. 

The same authors have suggested a new approach (P43) in 2023 where a DSL model was used for synthetic data generation. The DSL model was retrained on a quarterly basis on a training corpus that was composed of 100 days of production data that they had access to. Retraining the model on actual production data and regeneration of the whole synthetic dataset allowed to evolve the generated synthetic data on a quarterly basis. From the paper, it is not clear to us if essential attributes of data object instances were preserved, or if this evolution approach would allow them to be preserved.


\subsubsection{What methods exist for generating and/or evolving synthetic test data that imitate real-life data?}

Answering this RQ-facet was challenging, as there were not many of the selected publications that included a thorough description of the synthetic data that was generated. Based on the data extracted from our selected publications, we defined two categories:

\begin{itemize}
    \item \textbf{Real-life like data:} the publication included information about the generated synthetic data being real-life like.
    \item \textbf{NA:} the publication included no or not enough information about the generated synthetic data being real-life like.
\end{itemize}

As shown in Figure~\ref{fig1}, there were 24 studies among our 37 selected publications where we could identify that the generated synthetic data imitated real-life data. In this context, it is important to note that we were not in the position to evaluate the synthetic data ourselves and this decision was therefore made solely based on the information provided by the authors of the studies. In 13 studies, it was unclear if synthetic data was actually produced or if it was real-life like. We found that the most comprehensive descriptions of the synthetic data were in the studies where synthetic or semi-synthetic images were generated. These descriptions were created during the validation or evaluation of the approach.

\begin{figure*}
\centerline{\includegraphics[width=\textwidth]{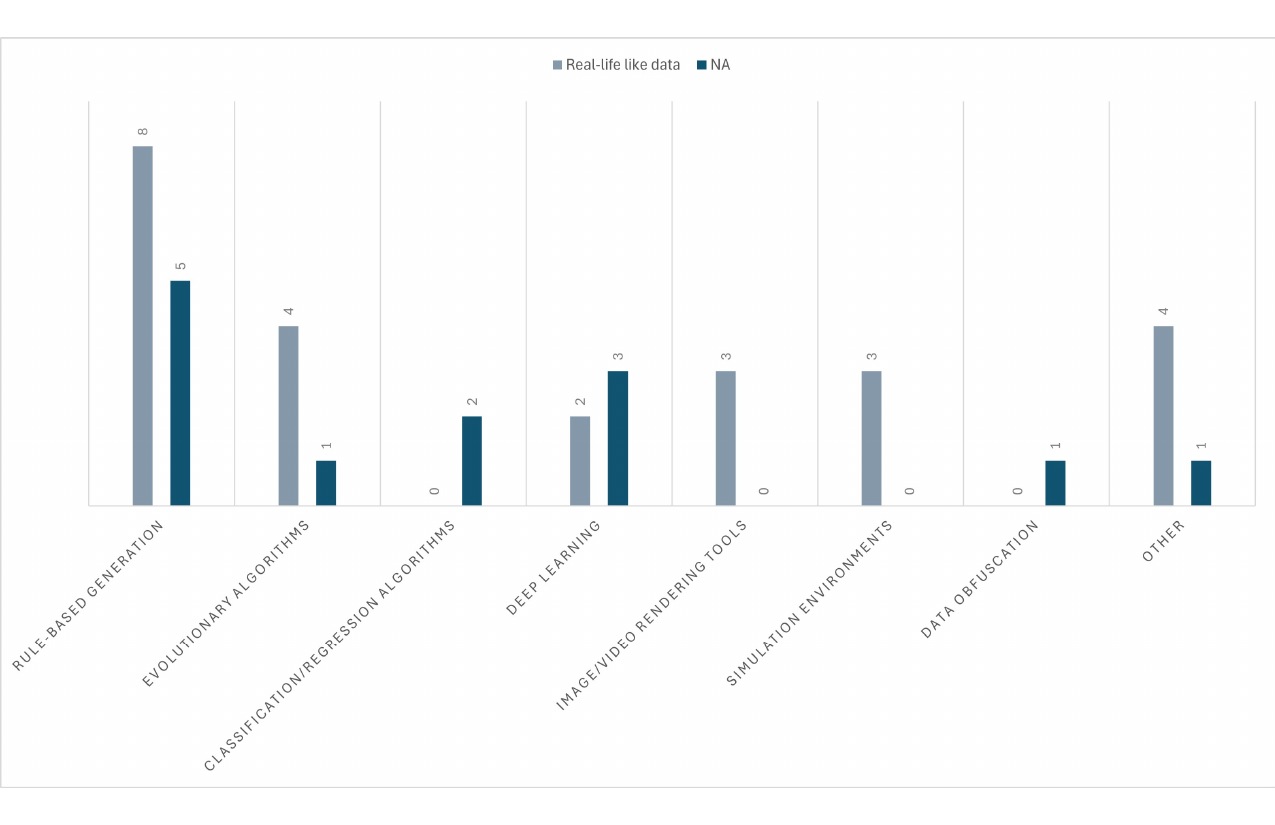}}
\caption{Type of approach - output data}
\label{fig2}
\end{figure*}

The two categories of generated synthetic data were distributed among the types of approaches shown in Figure~\ref{fig2}.

\subsubsection{What methods exist for generating and/or evolving synthetic test data without using real-life data as input?}

In the Title and Abstract Analysis and Full Text Analysis stages, we excluded all publications that suggested synthetic data generation approaches where real-life data is needed as input. In order to get a better understanding of what our 37 selected publications require as input when generating and/or evolving synthetic data, we classified the data item ``input data that is used as a starting point" using the following categories:

\begin{itemize}
    \item \textbf{No input data:} synthetic data is generated either according to specific rules defined by the user or with the help of any other means that do not require access to domain-specific data or source code.
    \item \textbf{Source code/SUT:} access to the source code of the SUT is required.
    \item \textbf{Existing test- or training datasets:} access to already existing test data of the SUT is required or publicly available test or training data is used, for example the Iris dataset from the UCI Machine Learning Repository~\citep{misc_iris_53}.
    \item \textbf{Random data:} data of pre-defined data type (e.g. string, integer) is created randomly without access to domain-specific data or source code. 
\end{itemize}

It is important to note that although no real-life data was used in the studies that relied on publicly available test or training datasets when validating or evaluating their approaches, it is not possible to apply them to real-life systems without gaining access to real-life data. Therefore, these approaches cannot be considered as an answer to the RQ-facet ``What methods exist for generating and/or evolving synthetic test data without using real-life data as input?"

The same applies to approaches that require access to the actual source code of the SUT. In our context, having access to the source code of the SUT cannot define the data that is coming in from external e-Government services. Therefore, in order to synthesize incoming data, these approaches would need access to real-life data, or they would need to be combined with another approach.

Based on these two considerations, our RQ-facet ``What methods exist for generating and/or evolving synthetic test data without using real-life data as input?" can only be answered with selected publications where either no input data or random data is required as input for the suggested data synthesis approaches.

\begin{figure*}
\centerline{\includegraphics[width=\textwidth]{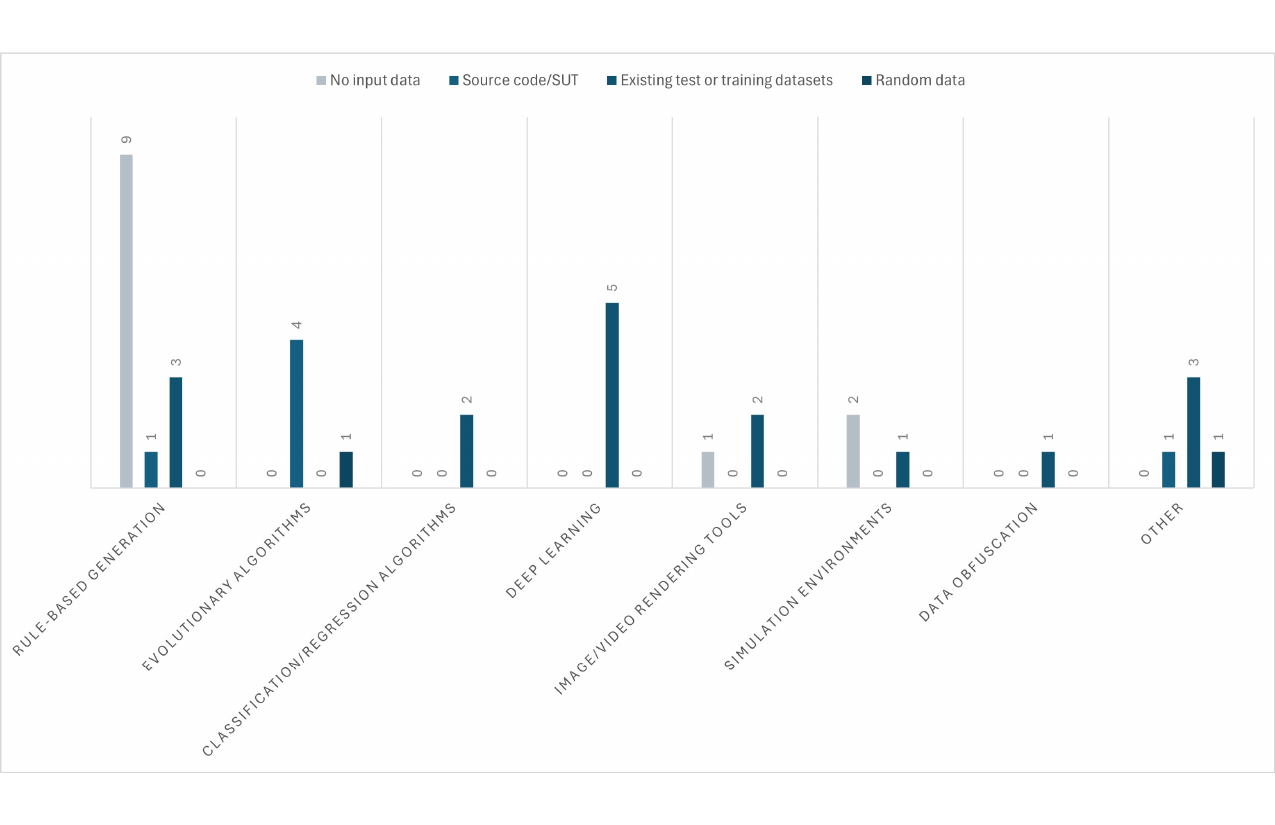}}
\caption{Type of approach - input data}
\label{fig3}
\end{figure*}

As shown in Figure~\ref{fig3}, this is an important limitation that almost entirely excludes the possibility of using Machine Learning approaches that have proven to be very effective for generating large amounts of realistic test data. The limitation is to some extent also relevant to using Large Language Models (LLMs). Therefore, among our 37 selected publications, we were able to identify only 14 approaches that provide an answer for the RQ-facet ``What methods exist for generating and/or evolving synthetic test data without using real-life data as input?''

\subsubsection{What methods exist for generating and evolving synthetic test data that imitate specific real-life data without using the respective real-life data as input?}

\begin{figure*}[ht]
\centerline{\includegraphics[width=\textwidth]{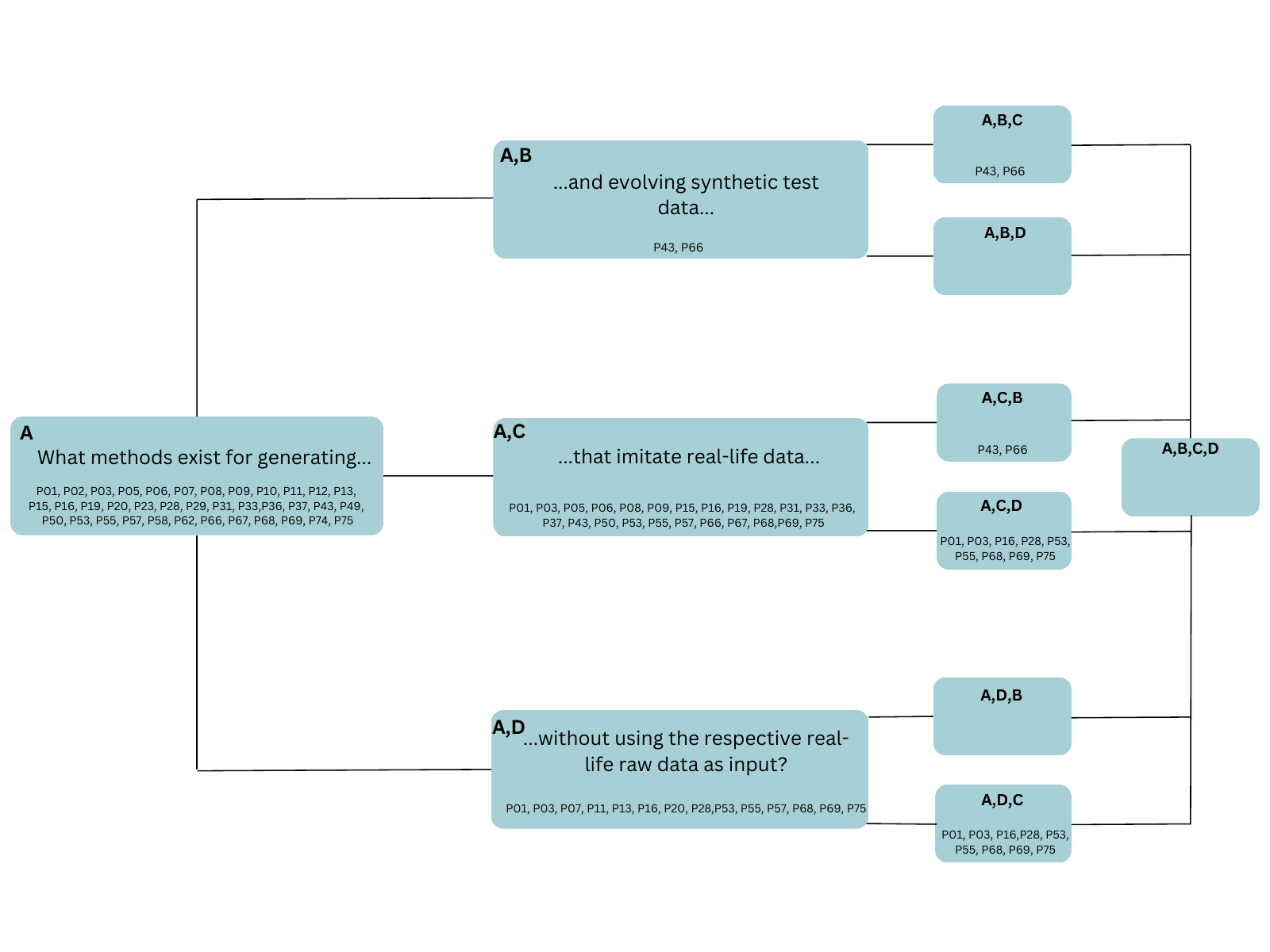}}
\caption{Answers to RQ-facets}
\label{fig4}
\end{figure*}

There was no approach suggested among our selected publications that provided an answer to all our four RQ-facets. There were however nine publications that came close by answering three out of our four RQ-facets, only missing one single solution for data evolution (see Figure~\ref{fig4}). 

The majority of the nine publications suggested a Rule-based Generation approach that required no input data (P01, P03, P28, P53, P69 and P75). One approach was based on the Successive Random Addition (SRA) method (P16) and the final two used Image/video rendering tools (P55) and Simulation environments (P68).

Four out of the nine publications stood out in our Quality Assessment where both (P28) or at least one of the researchers (P53, P55, P69) decided that all of our four Quality Assessment Criteria can be graded as ``Yes". 


Two out of the nine publications, P01 and P68, received poor evaluations. P01 was marked with a ``No” for every Quality Assessment Criteria by one of the researchers, and P68 was commented on as a ‘very poor paper’ by a researcher. The first publication (P01) proposed a testing technique that integrated an external test-case generator into a Property-Based Testing (PBT) tool in order to combine the features of two test-case generation strategies. Rules derived from WSDL descriptions are used as input for generating synthetic test data. The second publication with Quality Assessment results below average (2/4 from Researcher 1, and 1/4 from Researcher 2) and poor researcher comments (P68) was the only one in our 37 selected publications that was not published in peer-reviewed conferences or journals. It is a short paper consisting of twelve pages including References and Appendices, and it aims to use simulation for generating synthetic datasets with desired properties (number of examples, data changes events) for the evaluation of Multi-Target Regression and Multi-Label Classification methods.

All nine studies showed at least some validation and/or evaluation efforts with regard to the suggested approaches. There were publications where it was clear to the researchers if and how the validation and evaluation was done (e.g. P28, P53) as well as those where this information was presented rather vaguely (P68, P75). 

Six out of the nine studies used tool support (P01, P03, P28, P53, P55, P69) when generating synthetic data and for two studies the tool was accessible for the researchers at the time of this review (P28, P53).

\begin{itemize}
    \item \textbf{P01:} Rule-based generation of test cases. Might be usable to some extent when generating synthetic data based on WSDL descriptions. Limitations of the approach were described, and they were related to computational resources required. Poor Quality Assessment results from one Researcher (3/4 Researcher 1, 0/4 Researcher 2). Tool support was used when generating synthetic test data 
    (MoMuT\footnote{\url{https://momut.org/}} and FsCheck\footnote{\url{https://fscheck.github.io/FsCheck/}}), although the tools are available but there was no proper description of its replicability (conference paper). 
    
    \item \textbf{P03:} An integration of grammar-based testing in a framework for contract-based testing in PHP. No limitations described. Quality Assessment results 3/4 (Researcher 1), 1/4 (Researcher 2). Tool support was used (Praspel), but the website that was referenced in the publication as the location of the tool did not include it. Also, the tool does not seem to be active.
    
    \item \textbf{P16:} The Successive Random Addition (SRA) \citep{liu2004corrected} was used for synthetic data generation with the purpose of assessing spatio-temporal data quality. Quality Assessment results were average (2/4). No limitations and no tool support was described.
    
    \item \textbf{P28:} This publication suggested an approach based on UML\footnote{\url{https://www.omg.org/spec/UML/}} and OCL (Object Constrain Language) constraint solving that can generate synthetic data for system testing. OCL constrains solvers, an important part of model-driven engineering (MDE), allow us to find solutions to constrains expressed in OCL. The publication received the highest possible Quality Assessment score from both Researchers (4/4). Limitations were described, and they are imposed by the constraint solver that might not always be exhaustive and find a solution. Tool support was provided and the PLEDGE (PracticaL and Efficient Data GEnerator for UML)\footnote{\url{https://github.com/SNTSVV/PLEDGE}} tool was available on GitHub but it seems to be not active at the time of this review.
    
    \item \textbf{P53:} This publication proposed a new synthetic data generator that was able to generate three-way datasets with planted triclusters (where values are correlated across the three dimensions (observations $times$ features $times$ contexts)) where the user could define several properties regarding the dataset and the  planted solutions. The publication received maximal Quality Assessment results from one Researcher, 2/4 (Researcher 1) and 4/4 (Researcher 2). Limitations were discussed, and they are related to computational resources required for synthesizing data with this approach. The publication proposed tool support (G-Tric\footnote{\url{https://github.com/jplobo1313/G-Tric}}) and the tool was accessible at the time of this review.
    
    \item \textbf{P55:} In this publication, two existing tools were combined to synthetically generate facial data. The publication received maximal Quality Assessment results from one Researcher, 2/4 (Researcher 1) and 4/4 (Researcher 2). No limitations were described. Existing image rendering tools (iClone, Blender 3D) were used for synthetic data generation.
     
    \item \textbf{P68:} This publication uses simulation for generating synthetic datasets with desired properties (number of examples, data changes events) for the evaluation of Multi-Target Regression and Multi-Label Classification methods. Quality Assessment results are below average (2/4 Researcher 1, 1/4 Researcher 2). The limitation of this approach is the limited number of inputs and outputs. Tool support is not described.
    
    \item \textbf{P69:} This paper focuses on automatically generating valid test input data for jUnit tests based on the provided Design by Contract~\citep{Meyer97} specification and with help of mocking. The publication received half the marks in Quality Assessment results from one researcher and maximal from the other (2/4 Researcher 1 and 4/4 Researcher 2). Limitations are related to the performance of the suggested approach and the quality of generated synthetic data. Tool support is not described.
    
    \item \textbf{P75:} This publication describes a Rule-based software test data generation approach that is proposed as an alternative to either path/predicate analysis or random data generation. It was the oldest publication that was found with our search strings, as it was published in 1991. The publication received a Quality Assessment score of 3/4. No limitations and no tool support were described in this publication.
\end{itemize}

\begin{boxtext}%
{\noindent \textbf{KEY FINDINGS:    
    \begin{itemize}
          \item Although there are many synthetic data generation approaches available, the majority of them require real-life data when applied in a real-life context.           
          \item Approaches that generate synthetic data as well as evolve the generated synthetic data over time are very rare.         
          \item Current approaches are mainly based on other domains that cannot easily transferred, for example images, or UML and OCL which are becoming less supported by the Software Engineering community.
          \item Proven current technologies that worked in other domains including LLMs have yet to be explored in our context. 
    \end{itemize}
    }
}
\end{boxtext}

We publicly provide the spreadsheets used for the analysis of both steps, i.e., Title and Abstract Analysis worksheet as well as the data extraction sheet used for the Full Text Analysis on Figshare \footnote{\url{https://doi.org/10.6084/m9.figshare.25809616.v1}}. 


\section{Limitations and Threats to Validity}
\label{sec:ttv}

Inspired by the reflection of 
\cite{Verdecchia23,lago24}, we have identified the limitations, Threats to Validity (TTV), and the causal relationships between these two concepts in our specific study. Limitations are a broader concept, describing the scope of a study, focusing more on assessing the available options. On the other hand, TTV are the consequences of the choices made due to the specific Limitations that apply for our study.

From the variety of existing types of threats, we have defined and discussed those that are relevant to our research method. They are defined as follows:

\begin{itemize}
    \item \textbf{Internal Validity} examines whether an experimental treatment/condition makes a difference, and whether there is evidence to support a claim.
    \item \textbf{External Validity} concerns itself with whether the results can be generalized \citep{Ampatzoglou19}. 
\end{itemize}

The list of Limitations and the resulting TTV together with appropriate mitigation strategies (where applicable) is provided in Tables~\ref{tab:limitations} and~\ref{tab:mitigations}. The column ``Conclusion'' in Table~\ref{tab:mitigations} states if a TTV was accepted as it is or if actions were taken to reduce the effect of a TTV.

\begin{table*}
  \caption{Limitations and Threats to Validity}
  \label{tab:limitations}
  \begin{tabular*}{\textwidth}{@{\extracolsep\fill}p{5.5cm}p{5.5cm}p{5.5cm}@{\extracolsep\fill}}
    \toprule
    \textbf{Limitation} & \textbf{Threats to Internal Validity} & \textbf{Threats to External Validity}\\
    \midrule
    A limited number of databases were searched in this review. & \textbf{IV1:} The list of papers found may not be the full list of papers available in the world that answer our RQ & ---\\
    \hline
    The Exclusion and Inclusion of papers based on Title and Abstract was done mostly by the Principal Researcher. & \textbf{IV2:} Exclusion and inclusion results may be based on biased decisions. & ---\\
    \hline
    Data extraction is done by Researchers and not automated, therefore it is to some extent subjective. & \textbf{IV3:} Data extraction results may be biased and miss important data. & ---\\
    \hline
    The keywords describing the Population were not included in the Search String as they made the Search String too restrictive. & --- & \textbf{EV1:} The results of this review may not be transferrable to other similar Populations.\\
    \hline
    There are no standard criteria and metrics for Quality Assessment in secondary studies. & \textbf{IV4:} The choice and usage of Quality Criteria may be arbitrary. & ---\\
    \bottomrule
  \end{tabular*}
\end{table*}

\begin{table*}
  \caption{TTV Mitigation Strategies}
  \label{tab:mitigations}
  \begin{tabular*}{\textwidth}{@{\extracolsep\fill}p{1cm}p{8cm}p{8cm}@{\extracolsep\fill}}
    \toprule
    \textbf{TTV \#} & \textbf{Mitigation Strategy} & \textbf{Conclusion}\\
    \midrule
    \textbf{IV1} & The databases (IEEE Explore, ACM DL, and Scopus) that are the main sources for quality research in the field of Software Testing are selected. They contain papers that have been reviewed and therefore they have passed a preliminary quality inspection. 
    & The researchers accept the risk of missing some papers when not searching additional databases and grey literature because it is likely that most of these papers that are not published in journals will not meet the Exclusion-,  Inclusion-, Quality-, and Maturity criteria of our study.\\
    \hline
    \textbf{IV2} & To validate the decisions of the Principal Researcher, 20 papers were randomly selected for Abstract analysis and Exclusion/Inclusion of papers by second reviewers.
    
    The results of the experiment showed that three papers out of 20 were either excluded or not included by the Principal Researcher but they were included by a second reviewer.
    
    The full text of these three papers were then read by a second reviewer who decided that the stricter Exclusion/Inclusion strategy of the Principal Researcher is justified so that these papers should not be included.    
    & The researchers accept the risk of possibly not including up to 15\% of publications due to the stricter Exclusion/Inclusion strategy, because the stricter strategy was justified by the experiment.\\
    \hline
    \textbf{IV3} & To validate the data extraction results of the Principal Researcher, data from X papers was extracted by second reviewers.
    
    The extracted data for the data items \textbf{E-RQ-1}, \textbf{E-RQ-2}, \textbf{E-RQ-2.6}, \textbf{E-RQ-3}, \textbf{E-RQ-3.1} and \textbf{E-RQ-4} that were used for defining the relevance of each paper were therein after compared. There were nine principal discrepancies that were cleared, and the suggestions of the Principal Researcher were accepted.    
    & The Researchers accept the risk of possibly having biased data extraction results and missing important data because the experiment shows that the risk of failing to define the relevance of papers correctly based on the data extraction results of the Principal Researcher is low.\\
    \hline
    \textbf{IV4} & An approach proposed by 
    \cite{Dyba08} for assessing the quality of Qualitative Research by Principles of Good Practice for conducting Empirical Research in software engineering were customized and implemented.  
    & The Researchers have taken action to reduce the effect of this TTV.\\ 
     \hline
     \textbf{EV1} & Initial testing was done with trial test strings. The results of these tests showed the need to remove the keywords for Population from the search string because they made the search string too restrictive.
     This RQ facet was therefore included on the Exclusion/Inclusion criteria, as well as in the process of defining the relevance of papers.
    & The Researchers have eliminated this TTV by defining the Population in the Exclusion and Inclusion Criteria.\\ 
    \bottomrule
  \end{tabular*}
\end{table*}

\section{Discussion}
\label{sec:discussion}

As stated previously, our study aims to identify existing synthetic test data generation approaches that can be used in real-life context without having access to actual real-life data. In addition, we are interested in the ability of these approaches to evolve the generated synthetic data. Considering that our research question ``What methods exist for generating and evolving synthetic test data that imitate real-life data without using the respective real-life data as input?'' is quite restrictive, it is not surprising that our set of selected publications includes only 37 studies. Considering that software testing is not a new discipline and the need for test data has been there for decades, it is interesting that the majority of our selected publications (25) are published after 2015. It might be related to the fact that we were specifically looking for approaches where real-life data is not used in any step of the process. It is also quite recently that following the GDPR \citep{GDPR} that dates back to 2016 and entered into force in 2018, many countries in the world are enforcing stricter personal data protection laws. This has made access to real-life data increasingly complicated in the field of Software Testing. 

On the other hand, it shows that it is important to find efficient methods for creating fully synthetic test data without having access to actual real-life data.

There are two studies among our 37 selected publications that suggest a synthetic data generation approach as well as a synthetic data evolution approach. Surprisingly, both studies share three authors from the University of Oslo, Norway, who are working in cooperation with Testify AS for a goal very similar to ours, i.e., generating and evolving synthetic data for testing Digital Government Services. The authors start out without using actual real-life data from the Norwegian National Registry in 2019 and suggest Multi-layer Recurrent Neural Networks that are trained on synthetic test data. They also admit that the statistical characteristics of their training data are different from the actual data that is processed by the Norwegian National Registry. Four years later, in 2023, the same authors have switched over to using actual real-life data for training their suggested DSL model. This sequence of events really illustrates the complex challenge of generating synthetic data without using actual real-life data as input, as well as the importance of creating synthetic data that closely resembles real-life data. It also demonstrates that generating and evolving synthetic data for testing e-Government services is an important topic relevant in many countries where these services are widely used. Not only that, considering that many governments are on the digitization path, the number of these countries is likely to increase in the near future.

Although our strict Inclusion Criteria helped us to select only studies where no real-life data is used as input when generating synthetic data, answering our RQ-facet concerning input data forces us to really think about how approaches that are validated or evaluated only with test or training datasets, source code or SUT could really be applied in the context where no real-life data is available and no access to any source code or SUT is granted for security reasons. 
These restrictions cover other domains that are different from ours, such as images, but as we also were searching for techniques that can be applied to our domain, we were more open-minded in that respect.
The test or training datasets, source code, or SUT that are used for validation and evaluation in these studies cannot help us in real-life context, only real-life data can. Therefore, we consider these types of input data to be equivalent to real-life data in our specific context.

Among our 37 selected publications, none fully address our research question RQ. Nevertheless, there are nine studies that do not include synthetic data evolution ability, but provide an answer for all other three out of four RQ-facets. Of these nine studies, there are two that stand out as being of similar context to ours, showing great Quality Assessment results and Researcher comments, discussing the limitations of their approaches, and providing tool support (P28, P53). These two approaches are good candidates for future work.

The publications selected were also applied to different domains such as code or images that are not easily translated into our domain, data related to individuals that evolves over time, and life events. We were hoping that although the domain could differ the techniques could still be applied.

We need a metric(s) that tells us about the quality of the data if the evolution of the metric(s) follows the same value(s). These metrics would need to compare datasets, how similar two datasets are, and using these metrics along the time, if their differences are maintained.


\section{Conclusions and Future Work}
\label{sec:conclusions}

Digital government or e-Government solutions are implemented and used in many of the countries in the world in different degrees. Decentralization of e-Government solutions together with strict personal data protection laws poses new challenges to software testers testing these solutions, as test data received from one e-Government entity must be compatible with test data received from another e-Government entity. Real-life data processed by all e-Government entities may not be accessible at all, even for the generation of synthetic data. 

We conducted a literature review to identify existing approaches that can be used for generating real-life like synthetic data without using real-life data as input. We were also interested in finding out if any of the identified approaches is also able to evolve the generated synthetic data.

We found that although there are very many synthetic data generation approaches available, the majority of them require real-life data when applied in a real-life context. Even worse: in our case, approaches that generate synthetic data as well as evolve the generated synthetic data over time are very rare. 

There were nine primary studies among our selected publications that suggested an approach that can be used for creating real-life-like synthetic data without using real-life data in the process. None of them covered the synthetic data evolution. 

Our future work includes investigating the approaches suggested in two of these primary studies more thoroughly, as they could potentially be developed further and combined with a synthetic data evolution) method. Another possibility is the study of metrics to compare and analyse the quality of datasets used for testing and their evolution.








\bibliography{references}


\appendix

\bmsection{Quality Assessment}
\label{app2}

According to 
\cite{Kitchenham07}, it is critical to assess the quality of primary studies, in addition to using general inclusion and exclusion criteria. Quality assessment (QA) of studies remains a challenging task despite the variety of available QA instruments and practices \citep{Yang21}. 

For our study, we have implemented a customized version of the quality assessment criteria suggested by 
\cite{Dyba08}. These criteria, listed in Table~\ref{tab:quality}, are proposed for assessing the quality of qualitative research by principles of good practice for conducting empirical research in software engineering. From the original paper, we have excluded the criteria that were not relevant to our study (e.g. criteria that were explicitly aimed at research that involves participants). Quality criteria that were applicable for our study have been customized to the conditions of our research. Each of the criteria must be graded on a “yes” or “no” scale, whereby “no” also represents “not applicable”. 

\begin{table*}[ht]
  \caption{Quality Assessment Criteria}
  \label{tab:quality}
  \begin{tabular*}{\textwidth}{@{\extracolsep\fill}lp{17cm}@{\extracolsep\fill}}
    \toprule
    \textbf{ID} & \textbf{Quality Assessment Criteria}\\
    \midrule
    \textbf{QA-1} & Is there a clear statement of the aims of the research?\\
    \textbf{QA-2} & Is there an adequate description of the methodology in which the research was carried out?\\
    \textbf{QA-3} & Is there a clear statement of findings?\\
    \textbf{QA-4} & Is the approach valuable for research or practice?\\
    \bottomrule
  \end{tabular*}
\end{table*}

The quality assessment form presented in Table~\ref{tab:quality_assessment} was used by researchers of this survey. The quality assessment form includes an odd number of screening questions for every quality assessment criterion. If the majority of screening questions for a criterion are answered with ``yes'' then the final grade for the same criterion is ``yes''. Otherwise, the final grade is ``no''.

\begin{table*}
  \caption{Quality Assessment Form}
  \label{tab:quality_assessment}
  \begin{tabular*}{\textwidth}{@{\extracolsep\fill}lp{15cm}l@{\extracolsep\fill}}
    \toprule
    \textbf{ID} & \textbf{Screening questions} & \textbf{}\\
    \midrule
    \textbf{QA-1} & \textbf{Is there a clear statement of the aims of the research?}
    
    Consider:

    \begin{enumerate}
    \item Have the authors described the research gap in previous work related to this research (e.g. reference to previous papers of the authors, literature review, lack of related work, etc)? Is there a rationale for why the study was undertaken? 
    \end{enumerate}
    & Yes/No\\
    \hline
    \textbf{QA-2} & \textbf{Is there an adequate description of the methodology of the research?}
    
    Consider:
    \begin{enumerate}
    \item Have the authors described the research methodology in a way that allows the study to be repeated (e.g. detailed process descriptions are described, research data is available, etc)? 
    
    \item Are the limitations of the study discussed explicitly?
    
    \item Is the context in which the research was constructed precise? 
    \end{enumerate}
    & Yes/No\\
    \hline
    \textbf{QA-3} & \textbf{Is there a clear statement of study outcomes/findings?}
    
    Consider:
    \begin{enumerate}
    \item Are the findings explicit (e.g. magnitude of effect)?
    
    \item Has an adequate discussion of the evidence, both for and against the researcher’s arguments, been demonstrated?
    
    \item Has the researcher discussed the credibility of their findings?
    
    \item Are the findings discussed in relation to the original research questions?
    
    \item Are the conclusions justified by the results? 
    \end{enumerate}
    & Yes/No\\
    \hline
    \textbf{QA-4} & \textbf{Does the study describe the value of the research outcome for research or practice?}
    
    Consider:
    \begin{enumerate}
    \item Does the researcher discuss the contribution the approach makes to existing knowledge or understanding (e.g. do they consider the findings in relation to current practice or relevant research-based literature)?
    
    \item Does the research identify new areas in which research is necessary?
    
    \item Does the researcher discuss whether or how the findings can be transferred to other populations, or consider other ways in which the research can be used? 
    \end{enumerate}
    & Yes/No\\
    \bottomrule
  \end{tabular*}
\end{table*}

In this work, the work was divided such that at least two researchers assess every paper according to the quality assessment form. Should they come to a different conclusion regarding a specific criterion, they discuss their assessment results in detail and try to decide on the most appropriate conclusion. Should that not be possible, a third researcher decides the grade of the specific quality criterion.


\bmsection{Demographic analysis}\label{app3}

\begin{figure*}
    \centering
    \includegraphics[width=1.0\textwidth]
    {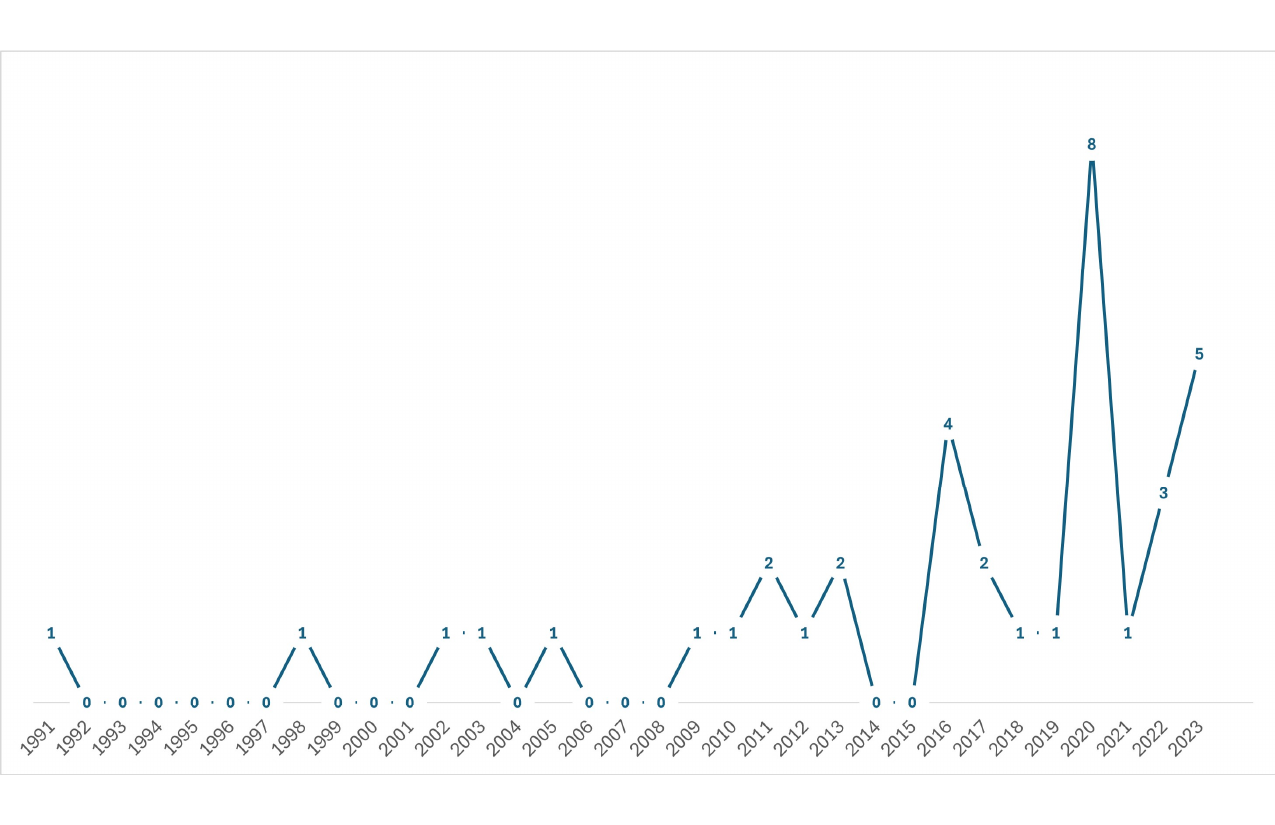}
    \caption{Demographics - timeline}
    \label{fig:demographics}
\end{figure*}

In the demographic analysis, the 37 selected findings are characterized based on their type of study, geographical location and affiliation of authors. As shown in Figure~\ref{fig:demographics}, all selected findings were published between 1991 and 2023, with the majority of them (25) published after 2015.
 
\begin{figure*}
    \centering
    \includegraphics[width=1.0\textwidth]
    {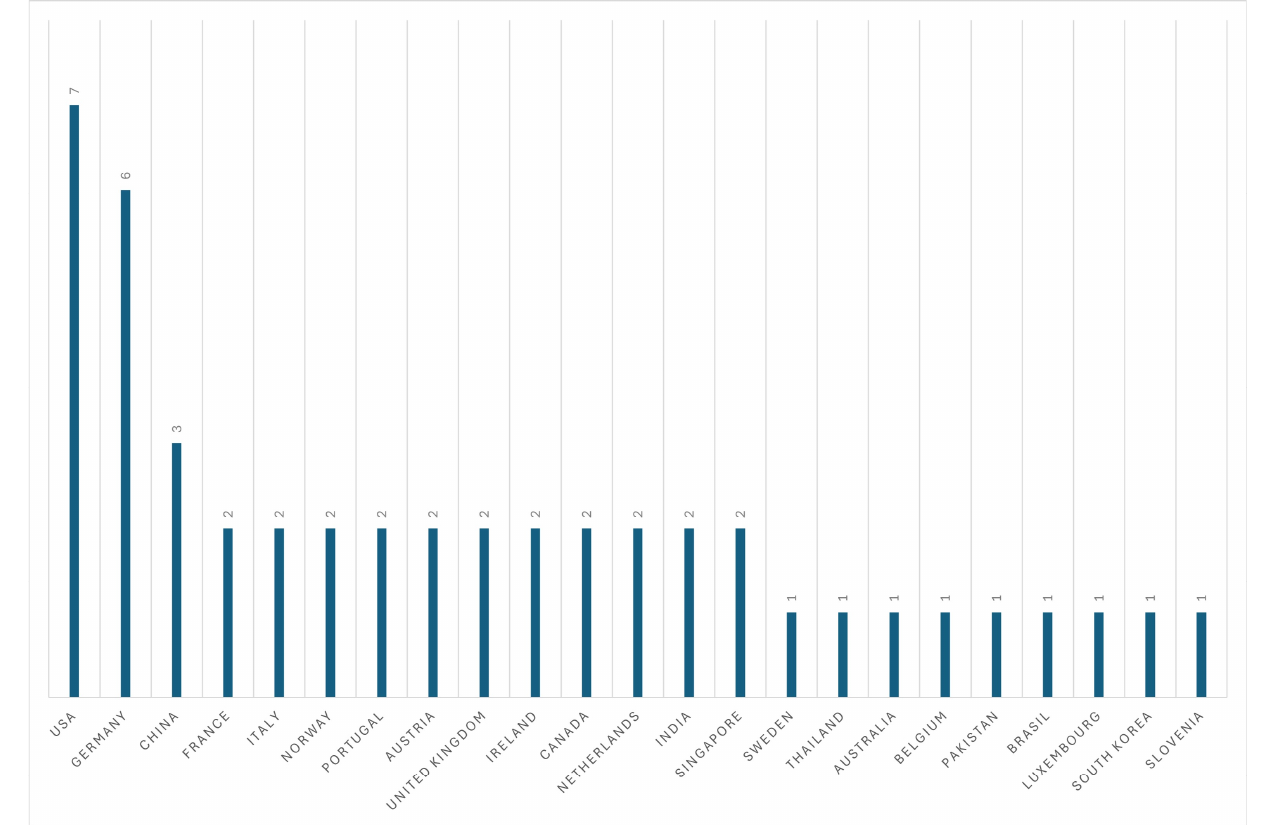}
    \caption{Demographics - geographical location}
    \label{fig:geographicalLoc}
\end{figure*}

We have based the geographical distribution of selected publications on the affiliation of the authors. A total of 119 unique authors were identified from our selected 37 publications. Three of these authors, Chao Tan (University of Oslo, Oslo, Norway and Testify AS, Oslo, Norway), Razieh Behjati (Testify AS, Oslo, Norway) and Erik Arisholm (Testify AS, Oslo, Norway) were among our list of authors twice with two papers among our selected publications. Our 37 selected publications had authors from 23 countries. The USA and Germany stand out by having affiliations in respectively 7 and 6 of our selected publications. The number of affiliations for each of the 23 countries is shown in Figure~\ref{fig:geographicalLoc}.

We categorized our selected publications according to the type of study according as follows:
\begin{itemize}
    \item \textbf{Academic:} all authors are affiliated with a university or research institute.
    \item \textbf{Industry/Public:} all authors are affiliated with a company, government institution or state agency.
    \item \textbf{Mixed:} some authors have an academic affiliation and some have an industry/public affiliation.
\end{itemize}

\begin{figure*}
    \centering
    \includegraphics[width=1.0\textwidth]
    {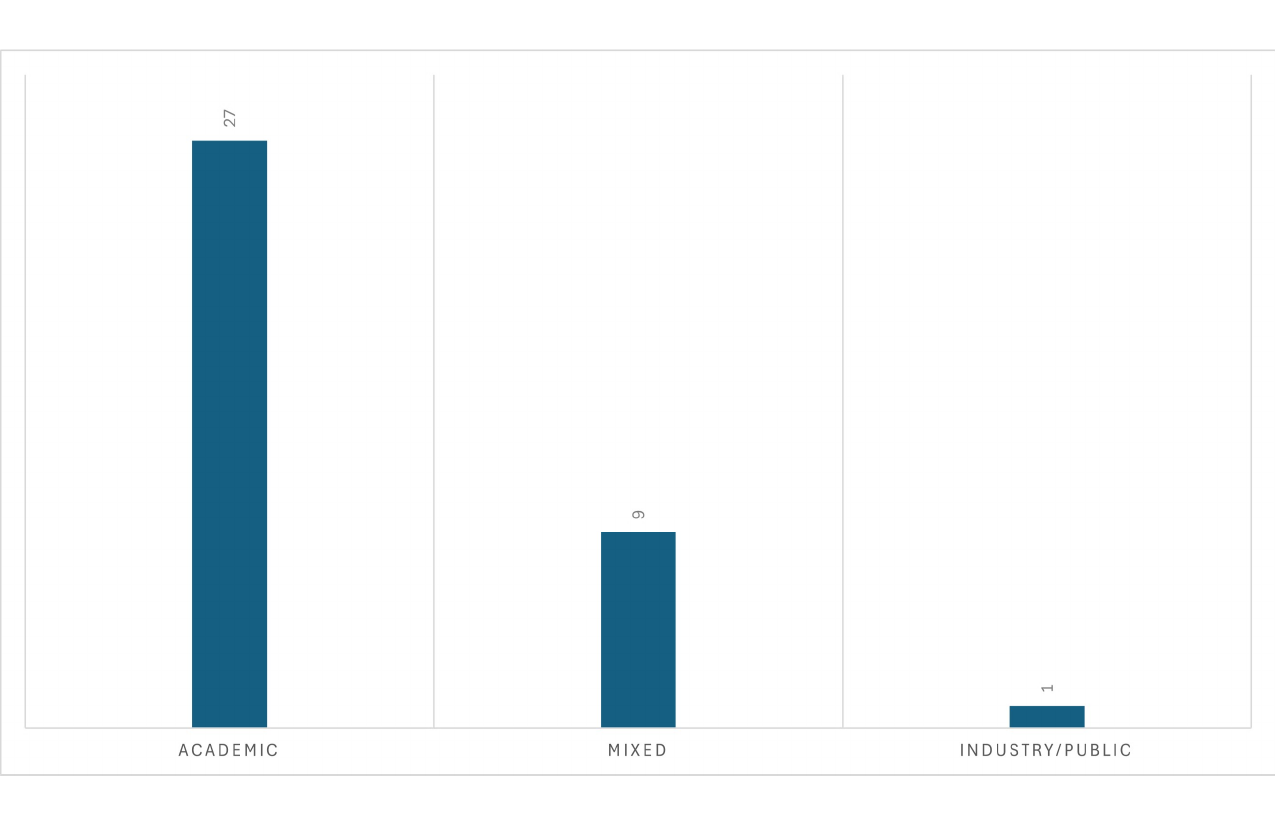}
    \caption{Demographics - type of study}
    \label{fig:demographicsTypeofStudy}
\end{figure*}

The majority of our selected publications (27) were purely academic based on the affiliations of authors. There were nine publications with mixed affiliations and one from industry. The distribution by type of study is shown on Figure~\ref{fig:demographicsTypeofStudy}.

\bmsection{Selected publications}

The 37 publications that were selected after Full Text Analysis are listed in Table~\ref{tab:selectedPub}.

\begin{table*}[ht !]
  \caption{Selected publications}
  \label{tab:selectedPub}
   \begin{tabular*}{\textwidth}{@{\extracolsep\fill}lp{10cm}l@{\extracolsep\fill}}%
    \toprule
    \textbf{ID} & \textbf{Name} & \textbf{DOI or URL}\\
    \midrule
    {P01} & Property-Based Testing with External Test-Case Generators & \href{https://doi.org/10.1109/ICSTW.2017.62}{10.1109/ICSTW.2017.62}\\
    P02 & Data coverage testing & \href{https://doi.org/10.1109/APSEC.2002.1183018}{10.1109/APSEC.2002.1183018}\\
    P03 & Grammar-Based Testing Using Realistic Domains in PHP & \href{https://doi.org/10.1109/ICST.2012.136}{10.1109/ICST.2012.136}\\
    P05 & An Experimental Tool for Search-Based Mutation Testing & \href{https://doi.org/10.1109/FIT.2018.00013}{10.1109/FIT.2018.00013}\\
    P06 & An Approach for Search Based Testing of Null Pointer Exceptions & \href{10.1109/ICST.2011.49}{https://doi.org/10.1109/ICST.2011.49}\\
    P07 & Research on Test Automation in the Field of Book Publishing Based on CNMARC Standards & \href{https://doi.org/10.1109/ICIECS.2009.5365826}{10.1109/ICIECS.2009.5365826}\\
    P08 & Evolutionary testing of unstructured programs in the presence of flag problems & \href{https://doi.org/10.1109/APSEC.2005.65}{10.1109/APSEC.2005.6}5\\
    P09 & Automatic test data generator: A tool based on search-based techniques & \href{https://doi.org/10.1109/ICRITO.2016.7785020}{10.1109/ICRITO.2016.7785020}\\
    P10 & Search Based Testing of Embedded Systems Implemented in IEC 61131-3: An Industrial Case Study & \href{https://doi.org/10.1109/ICSTW.2013.78}{10.1109/ICSTW.2013.78}\\
    P11 & Automatically generating realistic test input from web services & \href{https://doi.org/10.1109/SOSE.2011.6139088}{10.1109/SOSE.2011.6139088}\\
    P12 & Property-Driven Testing of Black-Box Functions & \href{https://doi.org/10.1145/3524482.3527657}{10.1145/3524482.3527657}\\
    P13 & A Strategy for using Genetic Algorithms to Automate Branch and Fault-based Testing & \href{https://doi.org/10.1093/comjnl/41.2.98}{10.1093/comjnl/41.2.98}\\
    P15 & DNN Analysis through Synthetic Data Variation & \href{https://doi.org/10.1145/3385958.3430479}{10.1145/3385958.3430479}\\
    P16 & A SMART Approach to Quality Assessment of Site-Based Spatio-Temporal Data & \href{https://doi.org/10.1145/2996913.2996932}{10.1145/2996913.2996932}\\
    P19 & Systematic Development of Data Mining-Based Data Quality Tools & \href{https://doi.org/10.5555/1315451.1315499}{10.5555/1315451.1315499}\\
    P20 & PLATOOL: A Functional Test Generation Tool for Mobile Applications & \href{https://doi.org/10.1145/3422392.3422508}{10.1145/3422392.3422508}\\
    P23 & On the Robustness of Aspect-Based Sentiment Analysis: Rethinking Model, Data, and Training & \href{https://doi.org/10.1145/3564281}{10.1145/3564281}\\
    P28 & Practical Constraint Solving for Generating System Test Data & \href{https://doi.org/10.1145/3381032}{10.1145/3381032}\\
    P29 & Fuzz Testing Based Data Augmentation to Improve Robustness of Deep Neural Networks & \href{https://doi.org/10.1145/3377811.3380415}{10.1145/3377811.3380415}\\
    P31 & CAD2Render: A Modular Toolkit for GPU-accelerated Photorealistic Synthetic Data Generation for the Manufacturing Industry & \href{https://doi.org/10.1109/WACVW58289.2023.00065}{10.1109/WACVW58289.2023.00065}\\
    P33 & Uncovering the Risks and Drawbacks Associated with the Use of Synthetic Data for Grammatical Error Correction & \href{https://doi.org/10.1109/ACCESS.2023.3310257}{10.1109/ACCESS.2023.3310257}\\
    P36 & GluGAN: Generating Personalized Glucose Time Series Using Generative Adversarial Networks & \href{https://doi.org/10.1109/JBHI.2023.3271615}{10.1109/JBHI.2023.3271615}\\
    P37 & Generation of meaningful synthetic sensor data — Evaluated with a reliable transferability methodology & \href{https://doi.org/10.1016/j.egyai.2023.100308}{10.1016/j.egyai.2023.100308}\\
    P43 & Enhancing Synthetic Test Data Generation with Language Models Using a More Expressive Domain-Specific Language & \href{https://doi.org/10.1007/978-3-031-43240-8\_2}{10.1007/978-3-031-43240-8\_2}\\
    P49 & Permutation-Invariant Tabular Data Synthesis & \href{https://doi.org/10.1109/BigData55660.2022.10020639}{10.1109/BigData55660.2022.10020639}\\
    P50 & Bayesian adversarial human motion synthesis & \href{https://doi.org/10.1109/CVPR42600.2020.00626}{10.1109/CVPR42600.2020.00626}\\
    P53 & G-Tric: generating three-way synthetic datasets with triclustering solutions & \href{https://doi.org/10.1186/s12859-020-03925-4}{10.1186/s12859-020-03925-4}\\
    P55 & Methodology for Building Synthetic Datasets with Virtual Humans & \href{https://doi.org/10.1109/ISSC49989.2020.9180188}{10.1109/ISSC49989.2020.9180188}\\
    P57 & SoccER: Computer graphics meets sports analytics for soccer event recognition & \href{https://doi.org/10.1016/j.softx.2020.100612}{10.1016/j.softx.2020.100612}\\
    P58 & Medical Time-Series Data Generation Using Generative Adversarial Networks & \href{https://doi.org/10.1007/978-3-030-59137-3\_34}{10.1007/978-3-030-59137-3\_34}\\
    P62 & Data Generators for Learning Systems Based on RBF Networks & \href{https://doi.org/10.1109/TNNLS.2015.2429711}{10.1109/TNNLS.2015.2429711}\\
    P66 & Synthetic test data generation using recurrent neural networks: A position paper & \href{https://doi.org/10.1109/RAISE.2019.00012}{10.1109/RAISE.2019.00012}\\
    P67 & Synthesis and evaluation of a mobile notification dataset & \href{https://doi.org/10.1145/3099023.3099046}{10.1145/3099023.3099046}\\
    P68 & First principle models based dataset generation for multi-target regression and multi-label classification evaluation & \href{https://ceur-ws.org/Vol-2069/STREAMEVOLV3.pdf}{https://ceur-ws.org/Vol-2069/STREAMEVOLV3.pdf}\\
    P69 & Automatically extracting mock object behavior from Design by Contract™ specification for test data generation & \href{https://doi.org/10.1145/1808266.1808273}{10.1145/1808266.1808273}\\
    P74 & SynConSMutate: Concolic testing of database applications via synthetic data guided by SQL mutants & \href{https://doi.org/10.1109/ITNG.2013.54}{10.1109/ITNG.2013.54}\\
    P75 & A Rule-Based Software Test Data Generator & \href{https://doi.org/10.1109/69.75894}{10.1109/69.75894}\\
    \bottomrule
  \end{tabular*}
\end{table*}






\end{document}